\documentclass[10pt,twocolumn,letterpaper]{article}

\usepackage[pagenumbers]{iccv} 
\usepackage{multirow}
\usepackage{makecell}
\usepackage{array}
\usepackage{marvosym}
\usepackage[accsupp]{axessibility}
\renewcommand{\thefootnote}{}
\definecolor{iccvblue}{rgb}{0.21,0.49,0.74}
\usepackage[pagebackref,breaklinks,colorlinks,allcolors=iccvblue]{hyperref}

\def\paperID{3699} 
\def\confName{ICCV}
\def\confYear{2025}

\title{Tune-Your-Style: Intensity-tunable 3D Style Transfer with Gaussian Splatting}

\author{
Yian Zhao\textsuperscript{1,3}$^\dag$ \ Rushi Ye\textsuperscript{1,3}$^\dag$ \ Ruochong Zheng\textsuperscript{1,3} \ Zesen Cheng\textsuperscript{1,3} \ Chaoran Feng\textsuperscript{1} \\
Jiashu Yang\textsuperscript{5} \ Pengchong Qiao\textsuperscript{1,3} \ Chang Liu\textsuperscript{4} \ Jie Chen\textsuperscript{1,2,3\ \Letter} \and
\small\textsuperscript{1}School of Electronic and Computer Engineering, Peking University, Shenzhen, China
 \quad
\small\textsuperscript{2}Pengcheng Laboratory, Shenzhen, China \\
\small\textsuperscript{3}AI for Science (AI4S)-Preferred Program, Peking University Shenzhen Graduate School, China \\
\small\textsuperscript{4}Department of Automation and BNRist, Tsinghua University, Beijing, China \quad
\small\textsuperscript{5}Dalian University of Technology, China
\vspace*{-4mm}
}

\begin{document}
\maketitle
\begin{abstract}
3D style transfer refers to the artistic stylization of 3D assets based on reference style images. 
Recently, 3DGS-based stylization methods have drawn considerable attention, primarily due to their markedly enhanced training and rendering speeds.
However, a vital challenge for 3D style transfer is to strike a balance between the content and the patterns and colors of the style.
Although the existing methods strive to achieve relatively balanced outcomes, the fixed-output paradigm struggles to adapt to the diverse content-style balance requirements from different users.
In this work, we introduce a creative intensity-tunable 3D style transfer paradigm, dubbed \textbf{Tune-Your-Style}, which allows users to flexibly adjust the style intensity injected into the scene to match their desired content-style balance, thus enhancing the customizability of 3D style transfer.
To achieve this goal, we first introduce Gaussian neurons to explicitly model the style intensity and parameterize a learnable style tuner to achieve intensity-tunable style injection.
To facilitate the learning of tunable stylization, we further propose the tunable stylization guidance, which obtains multi-view consistent stylized views from diffusion models through cross-view style alignment, and then employs a two-stage optimization strategy to provide stable and efficient guidance by modulating the balance between full-style guidance from the stylized views and zero-style guidance from the initial rendering.
Extensive experiments demonstrate that our method not only delivers visually appealing results, but also exhibits flexible customizability for 3D style transfer.
Project page is available at \href{https://zhao-yian.github.io/TuneStyle}{https://zhao-yian.github.io/TuneStyle}.
\end{abstract}
\footnote{\Letter\ Corresponding author. $^\dag$Co-first author.}
\vspace*{-5mm}
\section{Introduction}

\begin{figure}[ht]
\hsize=\linewidth
\centering
\includegraphics[width=\linewidth]{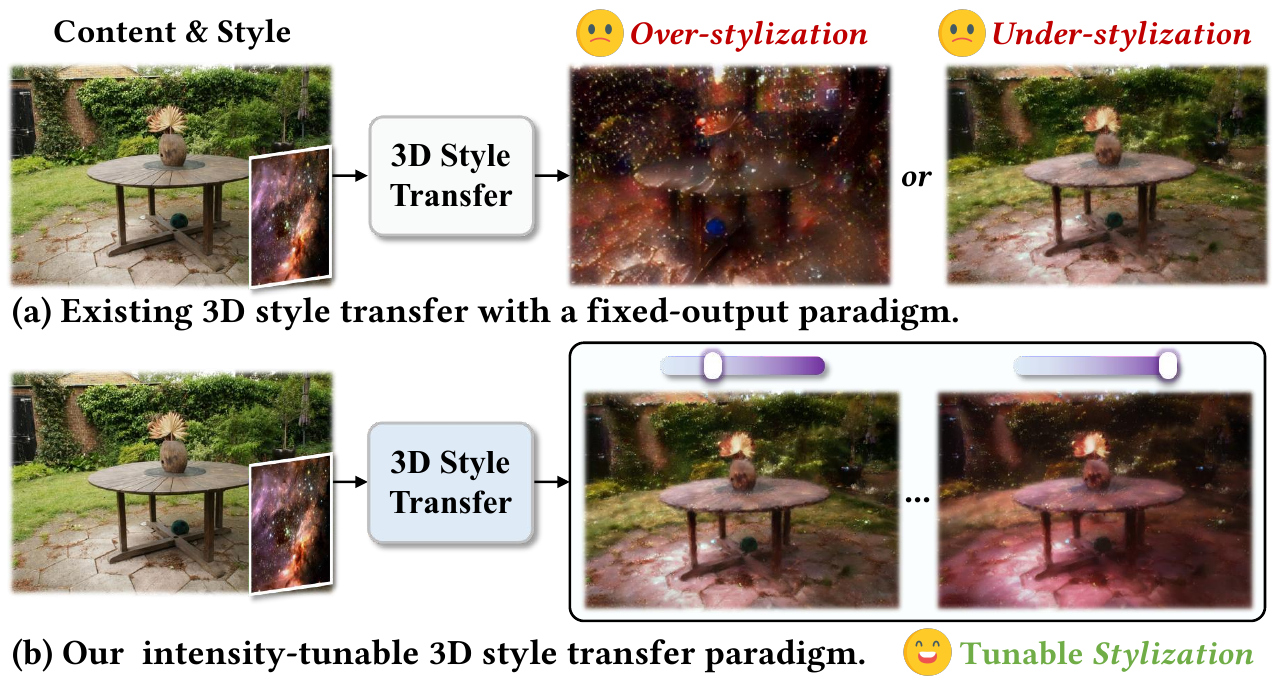}
\vspace{-7mm}
\caption{
\textbf{(a)} Existing fixed-output paradigm struggles to adapt to the diverse content-style balance requirements.
\textbf{(b)} Our intensity-tunable 3D style transfer paradigm enables users to flexibly adjust the style intensity to achieve the desired content-style balance.
}
\vspace{-6mm}
\label{fig:motivation}
\end{figure}

The technology of 3D style transfer holds promising application value in the fields of artistic creation, gaming, and entertainment. 
Conventional 3D style transfer approaches model scenes based on explicit representations (\eg, voxels, meshes, and point clouds), and the quality of their stylized results is limited by the fidelity of the appearance reconstruction. 
The advent of implicit neural representation, \ie, Neural Radiance Fields (NeRF)~\cite{mildenhall2021nerf}, marks a paradigm shift in the realm of 3D style transfer, inaugurating a new era of enhanced visual art effects. 
However, the complexity of implicit scene encoding results in time-consuming training and rendering for scene reconstruction, which hinders its practical application.

Recently, 3D Gaussian Splatting~(3DGS)~\cite{kerbl20233d} has attracted considerable attention for its advanced reconstruction quality and efficient rendering pipeline, offering a new technical pathway for 3D style transfer.
However, a vital challenge for 3D style transfer is to strike a balance between the content and the patterns and colors of the style.
Although existing methods~\cite{kovacs2024G,liu2024stylegaussian,zhao2025isegman} strive to achieve relatively balanced outcomes, the fixed-output paradigm is cumbersome to adapt to the diverse content-style balance requirements from different users, as presented in~\cref{fig:motivation}(a).

To address this dilemma, we present a creative intensity-tunable 3D style transfer paradigm \textbf{Tune-Your-Style}, which enables users to flexibly adjust the style intensity injected into the scene to match their desired content-style balance, substantially enhancing the customizability and practicality of 3D style transfer, as presented in~\cref{fig:motivation}(b).
Nevertheless, achieving this goal remains non-trivial due to the absence of explicit modeling and guidance for style intensity in the scene.
In response to this challenge, we first propose Intensity-tunable Style Injection (ISI), which introduces Gaussian neurons to explicitly model the style intensity and parameterizes a learnable style tuner to flexibly adjust the style intensity injected into the scene.
To facilitate the learning of the style intensity and tuner, we further propose Tunable Stylization Guidance (TSG).
TSG employs a 2D diffusion model to perform style transfer on rendered views and obtains multi-view consistent stylized results through cross-view style alignment. 
Subsequently, TSG adopts a two-stage optimization strategy to provide stable and efficient guidance by modulating the balance between full-style guidance from the stylized results and zero-style guidance from the initial rendering.

To verify the effectiveness of our approach, we performe extensive comparisons and visualizations across multiple scenes.
Specifically, we first conduct qualitative and quantitative comparisons with existing 3DGS-based and NeRF-based style transfer methods, and our approach exhibits markedly superior performance in terms of visual quality and quantitative metrics.
The widely collected user study results also highlight the distinct advantages of our method in user preference and visual appeal.
Furthermore, we present visualization results of two applications, \ie, intensity-tunable 3D stylization and multi-style combination, where our method exhibits impressive results and flexible customizability.

The main contributions can be summarized as: 
(\romannumeral1). We unfold a novel intensity-tunable 3D style transfer paradigm that addresses the challenge of the existing fixed-output paradigm being cumbersome to adapt to the diverse content-style balance requirements.
(\romannumeral2). We construct an explicit modeling of style intensity and parameterize a learnable style tuner to achieve intensity-tunable style injection.
(\romannumeral3). We design a tunable stylization guidance, which employs a two-stage optimization strategy to guide the learning of tunable stylization by modulating the balance between full-style guidance from the stylized results and zero-style guidance from the initial rendering.
(\romannumeral4). We conduct extensive comparisons and visualizations, where our method exhibits visually appealing results and flexible customizability.

\section{Related Work}

\subsection{Image/Video Style Transfer}
Image style transfer aims to create new artworks from real images based on artistic paintings. 
The pioneering work~\cite{Gatys_2016_CVPR} opens up a CNN-based optimization approach.
Building on this, subsequent studies~\cite{johnson2016perceptual,dumoulin2016learned,li2016precomputed,ulyanov2016texture,huang2017arbitrary,li2017universal,park2019arbitrary,liu2021adaattn,wu2021styleformer,Deng_2022_CVPR,Zhang_2023_CVPR,ye2023ip,wang2024instantstyle} have continuously innovated, striving for higher transfer efficiency and better stylized quality.
Video style transfer focuses on addressing the issues of inter-frame inconsistency and flickering artifacts in stylized videos, which are typically achieved through the design of temporal consistency constraints~\cite{chen2017coherent,huang2017real,ruder2018artistic,li2019learning,wang2020consistent,gao2020fast,deng2021arbitrary,wang2020consistent,yang2022vtoonify}.
Both image and video style transfer only support the stylization of 2D images and cannot be directly applied to 3D representations.

\subsection{3D Style Transfer}
3D style transfer aims to modify the appearance and geometry of a scene based on an image exemplar so that the rendering matches the desired style while maintaining content similarity and multi-view consistency, which is more challenging than image style transfer.
Existing mainstream methods can be categorized into two types: NeRF-based and 3DGS-based.

\noindent \textbf{NeRF-based.}
HyperNet~\cite{chiang2022stylizing} pioneers the 3D style transfer based on NeRF, achieving appearance stylization through a style hypernetwork.
Subsequent works~\cite{fan2022unified,huang2022stylizednerf,nguyen2022snerf,zhang2022arf} introduce various techniques to enhance the stylization quality, including the unified implicit neural stylization framework~\cite{fan2022unified}, the 2D-3D mutual learning strategy~\cite{huang2022stylizednerf}, the alternate training strategy~\cite{nguyen2022snerf}, and the nearest neighbor-based feature matching (NNFM) loss~\cite{zhang2022arf}.
More recently, \cite{haque2023instruct} and \cite{fujiwara2024style} achieve 3D style transfer from 2D diffusion priors, and \cite{wang2023nerf} enables text-driven 3D stylization based on CLIP~\cite{radford2021learning} features.
StyleRF~\cite{Liu_2023_CVPR} develops zero-shot 3D style transfer.
DeSRF~\cite{xu2023desrf} introduces a deformable module and dilated sampling for efficient style transfer.
CoARF~\cite{zhang2024coarf} provides fine-grained control over the resulting scenes. 
Although these methods achieve impressive results, they are limited by the time-consuming training of NeRF and only support a fixed-output paradigm, making it difficult to adapt to various content-style balance requirements.

\begin{figure*}[ht]
\hsize=\linewidth
\centering
\includegraphics[width=\linewidth]{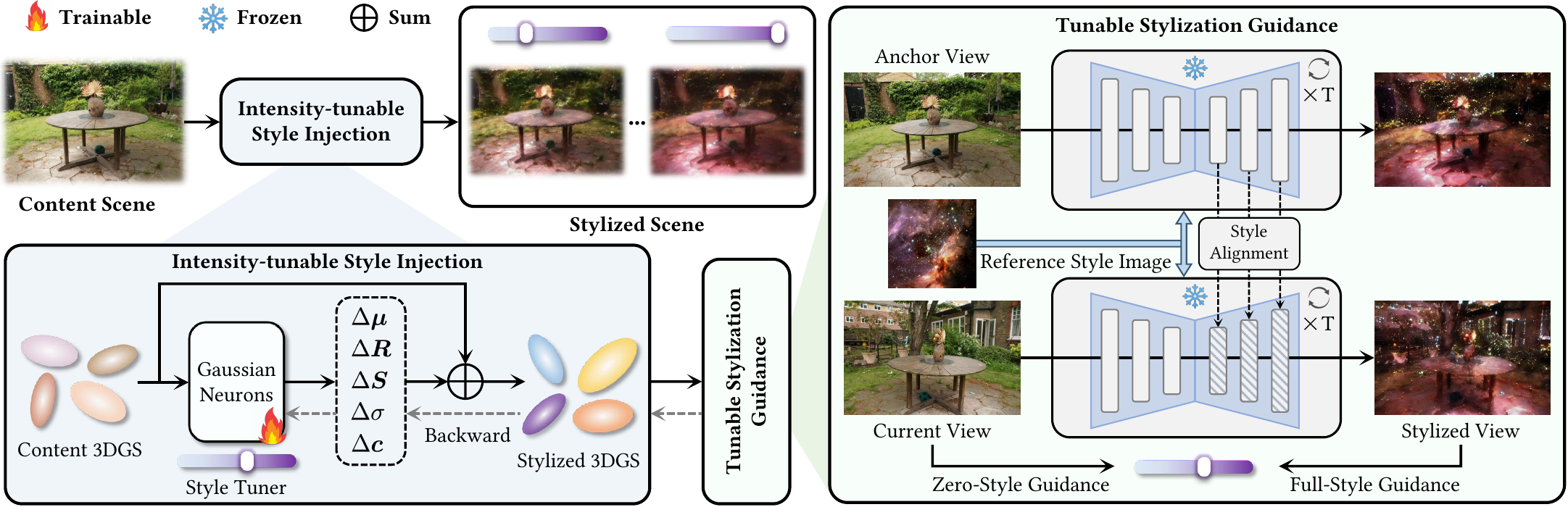}
\caption{
\textbf{Overall framework.}
Our method comprise two pivotal components, namely Intensity-tunable Style Injection (ISI) and Tunable Stylization Guidance (TSG).
ISI introduces Gaussian neurons to explicitly model style intensity and parameterizes a learnable style tuner, enabling users to flexibly adjust the style intensity injected into the scene.
To facilitate the learning of the style intensity and tuner, TSG first employs a diffusion model to perform style transfer on rendered views, and obtains multi-view consistent stylized results through cross-view style alignment. 
Then, TSG adopts a two-stage optimization strategy to achieve stable and efficient tunable stylization guidance, with full-style guidance in the stylized results and zero-style guidance in the initial rendering.
}
\vspace{-5mm}
\label{fig:overview}
\end{figure*}

\noindent \textbf{3DGS-based.}
Recently, 3DGS-based style transfer methods have gained attention for their efficient pipeline and high-quality scene fitting.
ReGS~\cite{mei2024reference} achieves reference-based scene stylization based on a content-aligned reference image.
StyleSplat~\cite{jain2024stylesplat} enables localized stylization of multiple objects within a scene.
StylizedGS~\cite{zhang2024stylizedgs} achieves adaptive control of multiple perceptual factors of stylized results.
StyleGaussian~\cite{liu2024stylegaussian} achieves instant style transfer using a KNN-based 3D CNN on embedded high-dimensional features.
InstantStyleGaussian~\cite{yu2024instantstylegaussian} leverages an improved iterative dataset update strategy to lift 2D diffusion priors for 3D style transfer.
G-Style~\cite{kovacs2024G} integrates preprocessing and multiple loss functions to obtain high-quality stylized results.
Although these 3DGS-based methods reduce training and rendering times compared to NeRF-based methods, they are also limited by the fixed-output paradigm.
In contrast, our method introduces a novel intensity-tunable style transfer paradigm and demonstrates visually appealing results, greatly enhancing the practicality of 3D style transfer.

\section{Method}

In this section, we elaborate on the proposed method, which comprises two pivotal components: Intensity-tunable Style Injection (ISI) and Tunable Stylization Guidance (TSG), as presented in~\cref{fig:overview}.
Specifically, ISI introduces Gaussian neurons to explicitly model the style intensity and parameterizes a learnable style tuner to flexibly control the style intensity injected into the scene.
TSG first employs a 2D diffusion model to perform style transfer on rendered views, and obtains multi-view consistent stylized results through cross-view style alignment. 
Subsequently, TSG adopts a two-stage optimization strategy to provide stable and efficient tunable stylization guidance by modulating the balance between full-style guidance from the stylized results and zero-style guidance from the initial rendering.
The details are as follows.

\subsection{Preliminary: 3D Gaussian Splatting}
3D Gaussian Splatting (3DGS)~\cite{kerbl20233d} introduces splatting-based rasterization and provides a real-time differentiable renderer.
Specifically, 3DGS represents a scene as a collection of 3D Gaussian primitives: $ \Theta = \{ (\boldsymbol{\mu}_i, \boldsymbol{S}_i, \boldsymbol{R}_i, \sigma_i, \boldsymbol{c}_i) \}_{i=1}^N $, where $N$ indicates the number of primitives, and each primitive consists of a position $\boldsymbol{\mu}_i \in \mathbb{R}^3$, a scaling factor $\boldsymbol{S}_i \in \mathbb{R}^3$, a rotation quaternion $\boldsymbol{R}_i \in \mathbb{R}^4$, an opacity $\sigma_i \in \mathbb{R}$, and a color $\boldsymbol{c}_i \in \mathbb{R}^k$~(where $k$ indicates the degrees of freedom). 
Practically, these 3D Gaussians can be effectively rendered to compute the color $\boldsymbol{C}$ of each pixel within the camera planes by blending $N$ ordered Gaussians overlapping the pixel.
For a more detailed introduction, please refer to Appendix D.

\subsection{Intensity-tunable Style Injection}

\noindent \textbf{Style Intensity Modeling.}
To model style intensity, we assign a learnable neuron to each Gaussian primitive to predict its attribute offsets, which serve as the quantitative representation of the style intensity.
Unlike previous methods~\cite{liu2024stylegaussian,yu2024instantstylegaussian} that typically only modify color attributes, we predict offsets for all attributes to finely model the style transfer of both geometry and appearance.
Given a scene $ \Theta $ and a reference style image $S_k$, the Gaussian neurons $\mathcal{G}$ predict the attribute offsets for all 3D Gaussians conditioned on $S_k$, as presented in~\cref{equ:neuron}.
\begin{equation}
\label{equ:neuron}
\mathcal{G}(S_k, \Theta) = \{ \Delta_k\boldsymbol{\mu}_i, \Delta_k\boldsymbol{S}_i, \Delta_k\boldsymbol{R}_i, \Delta_k\sigma_i, \Delta_k\boldsymbol{c}_i\ \}_{i=1}^N. 
\end{equation}
The stylized scene $\hat{\Theta}_{k}$ is calculated as in~\cref{equ:style_gs}.
\begin{equation}
\label{equ:style_gs}
\hat{\Theta}_{k} = \Theta + \mathcal{G}(S_k, \Theta).
\end{equation}
These Gaussian neurons parameterize the transformation field of the scene with respect to the reference style, thereby providing an explicit modeling of the style intensity.

\noindent \textbf{Tunable Style Injection.}
Based on the explicit modeling of style intensity, we further propose the tunable style injection to adjust the injected style intensity.
Specifically, we introduce a style tuner with two endpoints corresponding to zero style injection and full style injection.
To parameterize the style tuner, we define a staircase function $\mathcal{H}(\cdot)$ that quantizes and samples the continuous signal of the style tuner to obtain encodable discrete signals. The staircase function $\mathcal{H}(\cdot)$ is presented in~\cref{equ:gating_vector}.
\begin{equation}
\label{equ:gating_vector}
\mathcal{H}(\beta) = z - a, \ \mathcal{Q} \cdot z \le \beta \le \mathcal{Q} \cdot (z+1), 
\end{equation}
where $\beta$ $\in$ $[a, b]$ represents the input value of the style tuner, $a$ and $b$ denote two endpoints, typically 0\% and 100\%, $\mathcal{Q} = \frac{b-a}{Z}$ denotes the quantitative interval, $Z$ denotes the quantitative level, and $z$ $\in \{0, 1, \dots, Z-1\}$.
Subsequently, we construct a bijective mapping $f$ from discrete signals to learnable embeddings, \ie, $\mathcal{V}_\beta = f(\mathcal{H}(\beta))$.
In the context of $\mathcal{V}_\beta$, the stylized scene with the reference style $S_k$ is calculated by~\cref{equ:gating}.
\begin{equation}
\label{equ:gating}
    \hat{\Theta}_{k}^{\beta} = \Theta + \mathcal{V}_\beta \odot \mathcal{G}(S_k, \Theta),
\end{equation}
where $\odot$ represents the element-wise multiplication.

\noindent \textbf{3D Gaussian Filter.}
In addition, we introduce a 3D Gaussian filter to avoid assigning neurons to redundant 3D Gaussians, which would otherwise cause artifacts in the stylized scene.
Concretely, we observe that redundant 3D Gaussians commonly exist in scenes, which are not salient in the original rendering but produce non-negligible artifacts in the stylized rendering due to the reorganization of visibility and relative occlusion relationships of 3D Gaussians.
Consequently, the filter is responsible for removing those unimportant 3D Gaussians to reduce the impact on style transfer.
Inspired by~\cite{fan2023lightgaussian}, we calculate the importance score $\psi_i$ of each 3D Gaussian in rendering all training views, as calculated in~\cref{equ:filter}.
\begin{equation}
\label{equ:filter}
\psi_i = \sum^{C} \sum^{H\times W} \kappa(\Theta_i) \cdot \sigma_i \cdot \prod_{j=1}^{i-1} (1-\sigma_j),
\end{equation}
where $C$, $H$, and $W$ represents the number of training views, image height, and width, respectively. $\kappa(\Theta_i)$ represents whether the $i$-th 3D Gaussian is hit when calculating each pixel.
Intuitively, 3D Gaussians with fewer total hits, lower opacity, or occlusion contribute less to the rendering, and thus they are assigned lower importance scores.
The 3D Gaussians are sorted according to their importance scores, with the less significant ones being filtered out.

\subsection{Tunable Stylization Guidance}

\noindent \textbf{Diffusion-based Stylization Guidance.}
Existing methods~\cite{zhang2024stylizedgs,liu2024stylegaussian,kovacs2024G} typically align the VGG features~\cite{gatys2016image} of rendered views and style images via optimization, which involves time-consuming encoder-decoder training.
To avoid the costly training process, we leverage the 2D diffusion priors to efficiently guide the stylization of 3D scenes.
Formally, given the image-conditioned diffusion model $\mathcal{E}$, \eg, IP-Adapter~\cite{ye2023ip}, and a set of camera poses $\boldsymbol{\pi} = \{ \pi_v | v=1,2,\dots,C \}$, for any viewpoint $v$, we can render its view $\mathcal{I}_{v}$.
Subsequently, we conduct a diffusion denoising process on $\mathcal{I}_{v}$ conditioned on the reference style image $S_k$ to generate the stylized view $\mathcal{I}_{v}^k$.
The reference style image feature is selectively injected into the style-specific blocks of $\mathcal{E}$, which has proven effective in~\cite{wang2024instantstyle}.
These stylized views, which inherit the 2D diffusion priors, are utilized to guide the update direction of the 3D model.

\begin{figure*}[ht]
\hsize=\linewidth
\centering
\includegraphics[width=\linewidth]{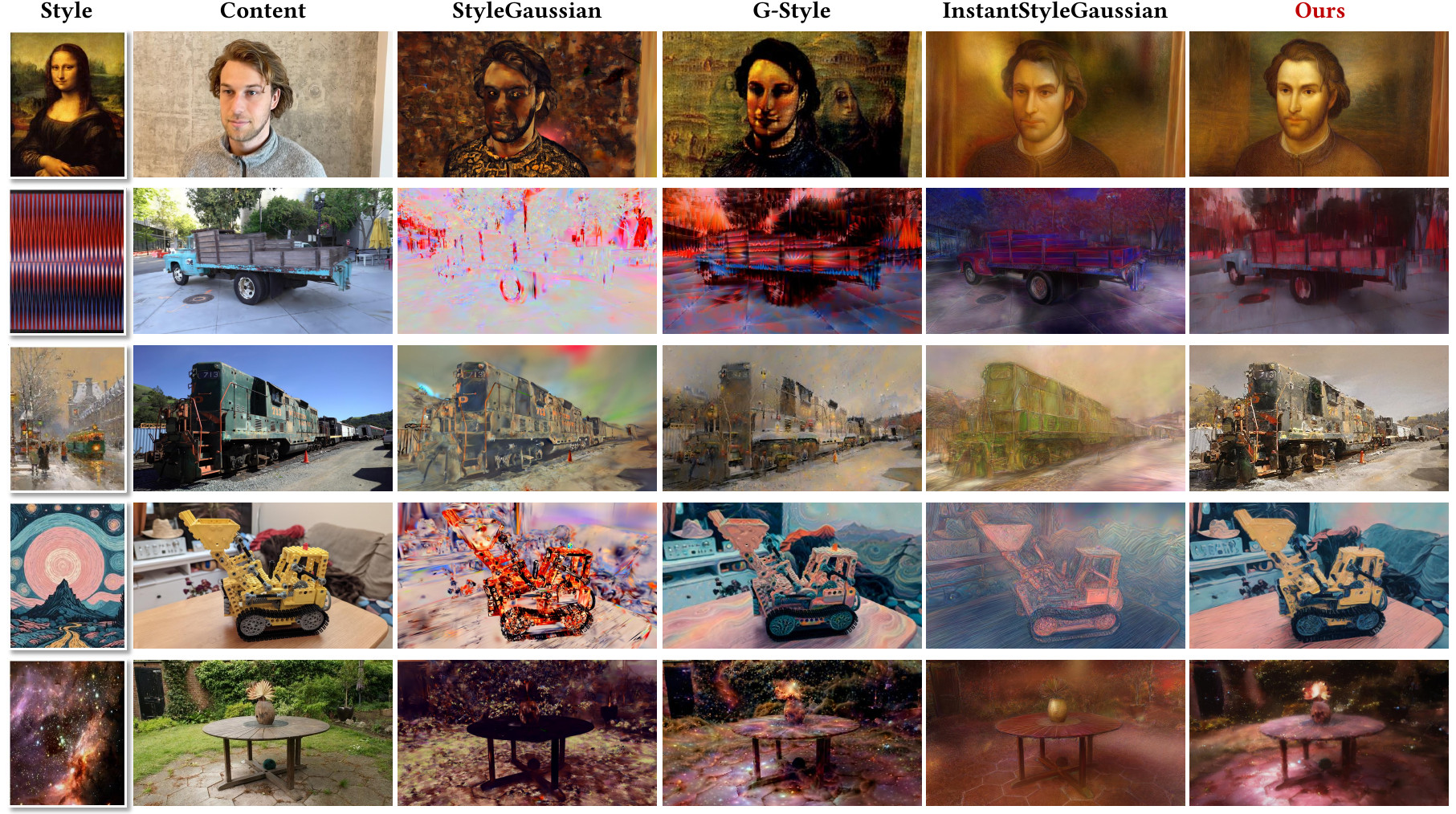}
\vspace{-6mm}
\caption{\textbf{Results of the qualitative comparison with 3DGS-based style transfer methods.} Our method exhibits clearer style textures, maintains more consistent color fidelity with the reference style, and produces fewer artifacts. Best viewed with zoom-in.
}
\vspace{-2mm}
\label{fig:comparison_gs}
\end{figure*}
\begin{table*}[t]
\centering
\renewcommand{\arraystretch}{1.15}
\setlength{\tabcolsep}{3.5mm}
\footnotesize
\begin{tabular*}{\linewidth}{l | cc | cc | c }
\toprule
\textbf{Method} & \textbf{Short-range Consistency} $\downarrow$ & \textbf{Long-range Consistency} $\downarrow$ & \textbf{CLIP$_{S}$} $\uparrow$ & \textbf{CLIP$_{S_{dir}}$} $\uparrow$ & \textbf{User Study} $\uparrow$ \\
\midrule
StyleGaussian~\cite{liu2024stylegaussian} & 0.067 / 0.070 &  0.126 / 0.108 & 0.2134 & 0.2223  & 2.79 $\pm$ 0.16 \\
G-Style~\cite{kovacs2024G} & 0.044 / 0.059 & 0.093 / 0.096 & 0.2406 & 0.2391 & 3.10 $\pm$ 0.40 \\
InstantStyleGaussian~\cite{yu2024instantstylegaussian} & 0.053 / 0.062 & 0.108 / 0.113 & 0.2204 & 0.2160 & 2.06 $\pm$ 0.22 \\
\textbf{Ours} & \textbf{0.033} / \textbf{0.035} & \textbf{0.062} / \textbf{0.067} & \textbf{0.2619} & \textbf{0.2881} & \textbf{3.97 $\pm$ 0.13} \\
\bottomrule
\end{tabular*}
\vspace{-2mm}
\caption{\textbf{Results of the quantitative comparison.} For the short- and long-range consistency, we present the results in the ``LPIPS / RMSE'' format.
CLIP$_{S}$ represents the CLIP similarity, and CLIP$_{S_{dir}}$ represents the CLIP directional similarity. For the user study, we report the average score and 95\% confidence interval.
}
\vspace{-4mm}
\label{tab:comparison_gs}
\end{table*}

\noindent \textbf{Cross-view Style Alignment.}
However, although the generated stylized views exhibit reasonable content and high-quality aesthetic texture, there is a distinct 3D inconsistency across different views, resulting in over-smoothing and blurring of 3D texture.
Existing 3D generation and editing methods typically apply the Score Distillation Sampling~(SDS) loss~\cite{poole2022dreamfusion} or construct an Iterative Dataset Updates~(IDU) process~\cite{haque2023instruct} to address this issue. 
Nevertheless, these strategies converge slowly due to the aggregation of inconsistent information across views, and they are especially deficient in handling stylized views with more intricate and detailed textures.

In response to this challenge, we propose the cross-view style alignment to achieve the multi-view consistent stylization guidance.
Specifically, we first randomly select an anchor view from the training views. Our goal is to align the style texture of the other views with that of the anchor view by injecting the features of the anchor view into the other views.
Inspired by~\cite{cao2023masactrl}, we inject the \textit{key} and \textit{value} features of the anchor view into the self-attention layers of other views during the diffusion process, and apply the mutual self-attention~\cite{cao2023masactrl} with the corresponding \textit{query} features to guide the stylization of other views.
However, we observe that directly applying mutual self-attention with the anchor view tends to distort the generation results of the current view due to the misalignment of features.
The degree of distortion escalates with the increase in view difference.
Consequently, we further apply cross-view feature matching to achieve content calibration.
To warp the content feature of the anchor view $\mathcal{I}_{v_a}$ to the current viewpoint $v_c$, we first back-project the feature into 3D space based on the camera pose of $\pi_{v_a}$ and the predicted depth map $\mathbf{d}_{v_a}$.
Subsequently, we re-project the spatial feature to the current viewpoint $v_c$ to obtain the warped content feature.
Finally, we concatenate the warped feature of the anchor view with the feature of the current view to serve as the \textit{key} and \textit{value} for mutual self-attention, which further improves the stability of content calibration.
Note that the feature injection is suggested to be applied to the latter layers of the model, which tend to determine the image texture~\cite{wang2024instantstyle,cao2023masactrl}. In our experiments, we apply it to the up\_blocks of U-Net~\cite{ronneberger2015u}.
The mutual self-attention is calculated in~\cref{equ:calibration}.
\begin{equation}
\label{equ:calibration}
\mathcal{A}_{v_c}^t = \text{softmax}(\frac{\mathit{Q}_{v_c}^t \cdot ([\mathit{K}_{v_a\rightarrow v_c}^t, \mathit{K}_{v_c}^t])^T}{\sqrt{d}}) \cdot [\mathit{V}_{v_a\rightarrow v_c}^t, \mathit{V}_{v_f}^t],
\end{equation}
where $\mathcal{A}_{v_c}^t$ represents the calibrated attention map of the current view $\mathcal{I}_{v_c}$ at time step $t$, $\mathit{K}_{v_a\rightarrow v_c}^t$ and $\mathit{V}_{v_a\rightarrow v_c}^t$ represent the warped \textit{key} and \textit{value} features of the anchor view, $[\cdot, \cdot]$ represents the concatenation.

\begin{figure*}[ht]
\hsize=\linewidth
\centering
\includegraphics[width=0.90\linewidth]{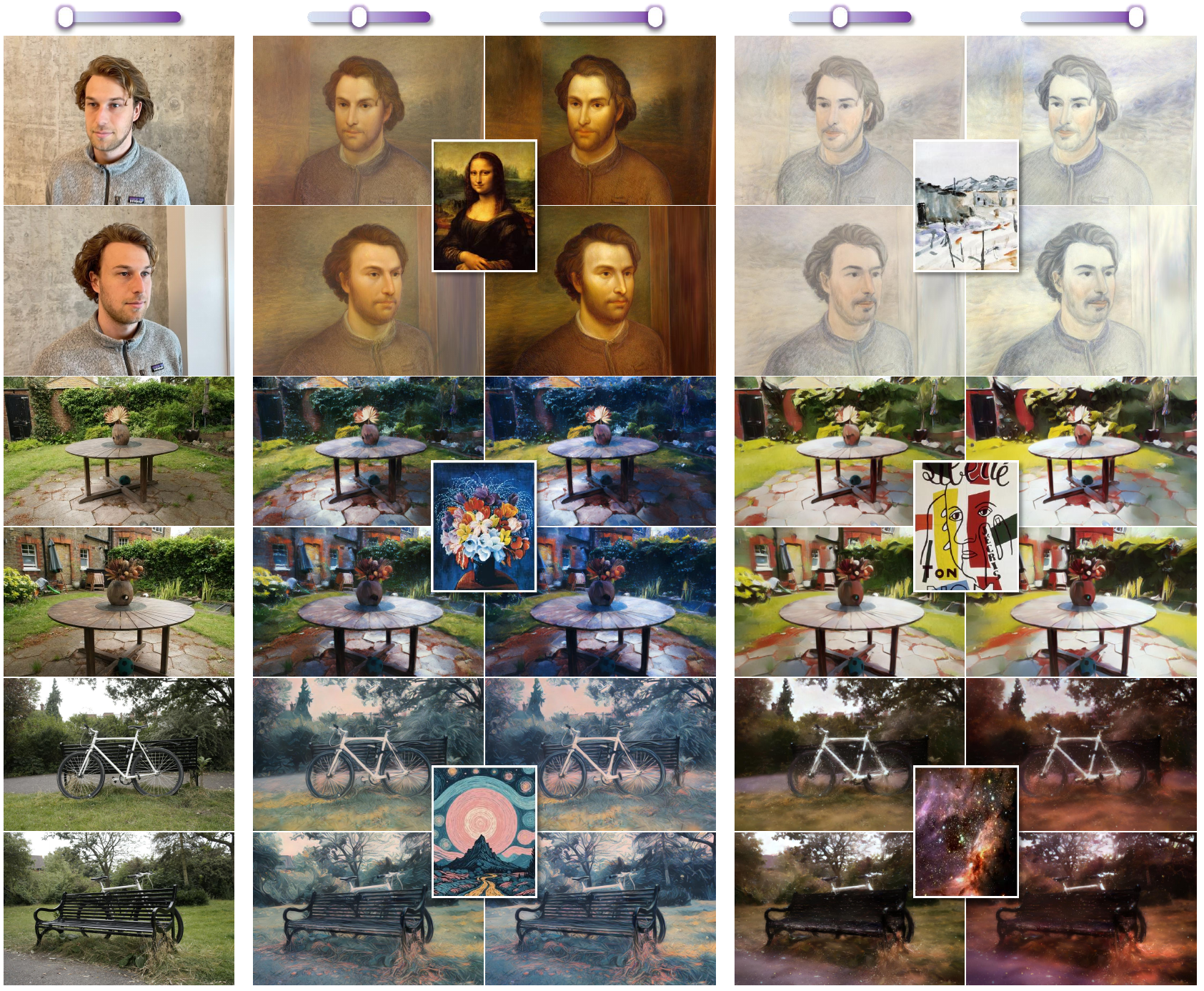}
\vspace{-1mm}
\caption{
\textbf{Results of intensity-tunable 3D stylization.}
The style tuner is marked above each column of rendered results, with the style injection intensity rising in proportion to the tuner value. Users can adjust the style tuner to their preferences to obtain desired results.
}
\vspace{-5mm}
\label{fig:injection}
\end{figure*}

\noindent \textbf{Two-stage Optimization for Tunable Stylization.}
Based on the multi-view consistent stylized results, we propose a two-stage optimization strategy for guiding the tunable stylization.
In the first stage, we perform the full-style guidance.
To ensure stable optimization in the early steps of training, the Gaussian neurons $\mathcal{G}$ are initialized with zeros and the learnable embeddings of the style tuner are initialized with ones, so that the initial predicted offsets are zero.
At this stage, the style tuner maintains a full offset position. Therefore, only the neurons $\mathcal{G}$ and the style tuner embedding corresponding to the full offset $\mathcal{V}_{full}$ are optimized.
Given the reference style image $S_k$, we calculate the $\mathcal{L}_1$ loss and the LPIPS~\cite{zhang2018unreasonable} loss as objective functions between the stylized view $\mathcal{I}_v^k$ and the rendered view $\mathcal{I}_v^{\tilde{t}_1}$, \textit{cf.}~\cref{equ:full}.
\begin{equation}
\label{equ:full}
\mathcal{L}_{full}^{s_1} = \mathcal{L}_1(\mathcal{I}_v^{\tilde{t}_1},\mathcal{I}_v^k) + \mathcal{L}_{lpips}(\mathcal{I}_v^{\tilde{t}_1},\mathcal{I}_v^k), 
\end{equation}
where $\tilde{t}_1$ represents the iteration step of the first stage, $\mathcal{L}_{full}^{s_1}$ represents the full-style guidance.
In the second stage, we perform the tunable stylization guidance.
At this stage, the neurons and the style tuner embedding of full offset are frozen, and only the other embeddings are optimized.
To provide tunable stylization guidance, we introduce zero-style guidance as an auxiliary.
Specifically, at each iteration $\tilde{t}_2$, we randomly sample the input of the style tuner $\beta_{\tilde{t}_2}$ (except for two endpoints) and calculate the corresponding embedding $\mathcal{V}_{\beta_{\tilde{t}_2}}$ according to~\cref{equ:gating_vector}.
Subsequently, we calculate the loss between the rendered view $\mathcal{I}_v^{\tilde{t}_2}$ and the original view $\mathcal{I}_v$ as the zero-style guidance, as well as between $\mathcal{I}_v^{\tilde{t}_2}$ and the stylized view $\mathcal{I}_v^k$ as the full-style guidance. 
The weighted sum of these two guidance, modulated by the style tuner value $\beta_{\tilde{t}_2}$, is employed as the tunable stylization guidance. 
For detailed formulas, refer to~\cref{equ:stage2_zero} and ~\cref{equ:stage2_full}.
\begin{equation}
\label{equ:stage2_zero}
\mathcal{L}_{zero} = \mathcal{L}_1(\mathcal{I}_v^{\tilde{t}_2},\mathcal{I}_v) + \mathcal{L}_{lpips}(\mathcal{I}_v^{\tilde{t}_2},\mathcal{I}_v),
\end{equation}
\begin{equation}
\label{equ:stage2_full}
\mathcal{L}_{full}^{s_2} = \mathcal{L}_1(\mathcal{I}_v^{\tilde{t}_2},\mathcal{I}_v^k) + \mathcal{L}_{lpips}(\mathcal{I}_v^{\tilde{t}_2},\mathcal{I}_v^k),
\end{equation}
where $\mathcal{L}_{zero}$ is the zero-style guidance, $\mathcal{L}_{full}^{s_2}$ is the second stage full-style guidance.
The tunable stylization guidance is calculated in \cref{equ:tunable_loss}.
\begin{equation}
\label{equ:tunable_loss}
\mathcal{L}_{tunable} = (1-\beta_{\tilde{t}_2}) \cdot \mathcal{L}_{zero} + \beta_{\tilde{t}_2} \cdot \mathcal{L}_{full}^{s_2}.
\end{equation}
\section{Experiments}
\label{sec:exp}

\begin{figure*}[ht]
\hsize=\linewidth
\centering
\includegraphics[width=0.98\linewidth]{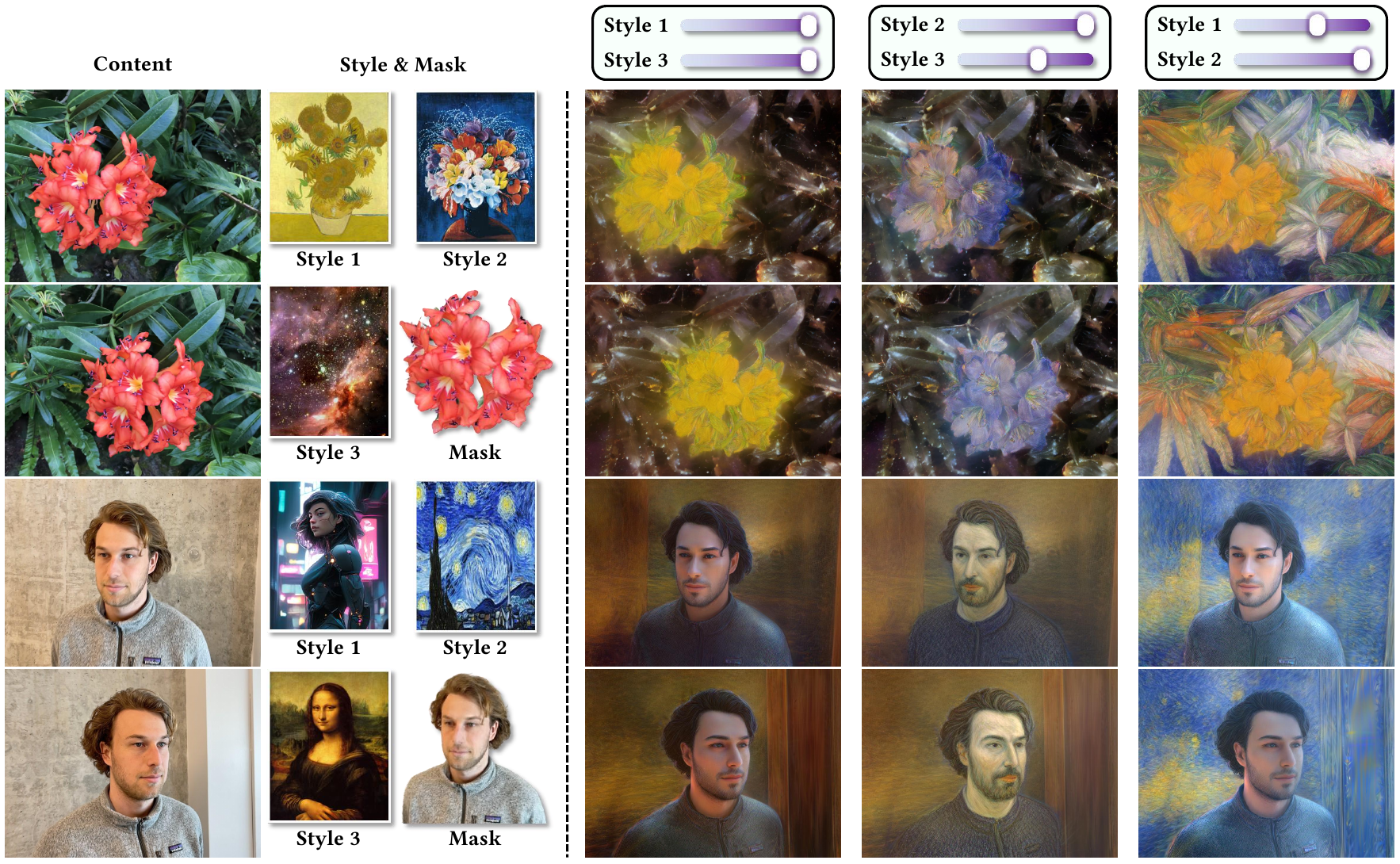}
\caption{
\textbf{Results of multi-style combination.}
We select two different styles to inject into the interior and exterior regions of the mask in each instance, and we can independently adjust the injection intensity of each style, thus controlling the dominant style of stylized scene.
}
\vspace{-5mm}
\label{fig:fusion}
\end{figure*}

\subsection{Implement Details}
In our experiments, all of the original 3D Gaussians are trained employing the method presented in~\cite{kerbl20233d}, with raw data from from Mip-NeRF~\cite{barron2022mip}, Tanks\&Temples~\cite{knapitsch2017tanks}, LLFF~\cite{mildenhall2019local}, and Instruct-N2N~\cite{haque2023instruct} datasets. 
We set the quantitative level $Z$ in the staircase function $\mathcal{H}(\cdot)$ of the style tuner to $10$, and the two endpoints $a$ and $b$ to 0\% and 100\%~\cite{zhao2024graco}, respectively.
We omit modeling the view-dependency of the color for simplicity and filter out the least important 50\% of the Gaussian primitives based on their importance scores.
The Gaussian neurons used for modeling style intensity are structured to match the number of parameters of the filtered Gaussian primitives.
Leveraging the highly optimized renderer of \cite{kerbl20233d}, these Gaussian neurons achieve efficient optimization while maintaining low memory consumption.
We utilize IP-Adapter-SDXL~\cite{ye2023ip} as the 2D diffusion model to provide stylization guidance, and inject the reference style image feature exclusively into the first up\_block following~\cite{wang2024instantstyle}.
The two-stage optimization of tunable stylization guidance comprises a total of 4000 steps, with each stage consisting of 2000 steps.
We use PyTorch for implementation and a single 32GB NVIDIA V100 GPU for training. The total training time for each scene is approximately 20 minutes.

\subsection{Main Results}
\noindent \textbf{Qualitative Comparison.}
We first conduct a qualitative comparison with existing 3DGS-based style transfer methods, as presented in~\cref{fig:comparison_gs}.
We compare our method with StyleGaussian~\cite{liu2024stylegaussian}, G-Style~\cite{kovacs2024G}, and InstantStyleGaussian~\cite{yu2024instantstylegaussian} on five commonly used scenes: face, truck, train, lego, and garden.
The stylization results of existing 3DGS-based methods still exhibit inadequacies in terms of content preservation and style texture details, and are also plagued by blurring and artifact noise.
In contrast, our method exhibits clearer style textures, maintains more consistent color fidelity with the reference style, and produces fewer artifacts.
Additionally, we also conduct comparisons with NeRF-based methods~\cite{huang2022stylizednerf,zhang2022arf,Liu_2023_CVPR}, with visualizations presented in Appendix A.
By leveraging the diffusion-based stylization prior, our method is able to produce more rational and visually appealing stylized outcomes.

\noindent \textbf{Quantitative Comparison.}
We further conduct a quantitative comparison of 3D style transfer quality with existing methods, \textit{cf.}~\cref{tab:comparison_gs}.
Specifically, we first calculate the long- and short-range multi-view consistency scores of the stylized scenes following common practice~\cite{huang2021learning,huang2022stylizednerf,liu2024stylegaussian}.
We also calculate the CLIP similarity and CLIP directional similarity scores to evaluate the fidelity to the reference style.
The former measures the cosine similarity of CLIP features between rendered views and text descriptions, while the latter measures the consistency of the change between rendered views with text descriptions.
Moreover, we conduct a user study to evaluate the user preference for the stylized results of different methods.
For multi-view consistency, we select 10 view pairs for each of the 5 scenes depicted in~\cref{fig:comparison_gs}. Among them, view pairs with an interval of 2 are utilized to calculate short-range consistency, while view pairs with an interval of 10 are utilized to calculate long-range consistency. 
We employ the same view warping algorithm as~\cite{liu2024stylegaussian}, \ie, optical flow~\cite{teed2020raft} estimation based on softmax splatting~\cite{niklaus2020softmax}, and calculate the LPIPS~\cite{zhang2018unreasonable} and RMSE scores for each view pair.
We calculate the average score of the five stylized scenes and present the results in the ``LPIPS / RMSE'' format.
Our approach exhibits optimal multi-view consistency.
To calculate CLIP similarity and directional similarity, we manually annotate the text descriptions for the original scenes and the stylized scenes. 
We employ CLIP ViT-L/14 as the feature extractor and also report the average score of the five scenes. Our method also demonstrates the strongest performance.
The detailed annotation information for these scenes can be found in Appendix C.1.
For the user study, we render 2 views for each scene and provide the original scene, the reference style image and the stylized scene to the participants.
We request participants to score on three dimensions: content consistency, style consistency, and visual appeal, using a 5-point scale.
We provide the average score and the 95\% confidence interval, and the results demonstrate that our method achieves the optimal user preferences.
The detailed evaluation criteria are provided in Appendix C.2.

\noindent \textbf{Intensity-tunable 3D Stylization.}
Subsequently, we present the visualization results of the intensity-tunable 3D stylization.
For each style, we present the results from two viewpoints, each incorporating two different style intensities, as shown in~\cref{fig:injection}.
The style tuner is marked above each column of rendered results, with the style intensity rising in proportion to the tuner.
Users can adjust the style tuner to their preferences to avoid over- or under-stylization, delivering flexible customizability with visually appealing results.
More results can be found in Appendix B.

\noindent \textbf{Multi-Style Combination.}
We also present the results of the multi-style combination.
For each scene, we employ SAM~\cite{kirillov2023segment} to segment multi-view images and subsequently obtain 3D masks following~\cite{chen2024gaussianeditor}. 
We assign three reference style images to each scene, from which we select two distinct styles to inject into the interior and exterior regions of the mask, respectively, as shown in~\cref{fig:fusion}.
Taking advantage of the intensity-tunable design, we can individually adjust the injection intensity of each style, thereby controlling the dominant style of the stylized scene.

\subsection{Ablation Study}

\begin{figure}[t]
\hsize=\linewidth
\centering
\includegraphics[width=\linewidth]{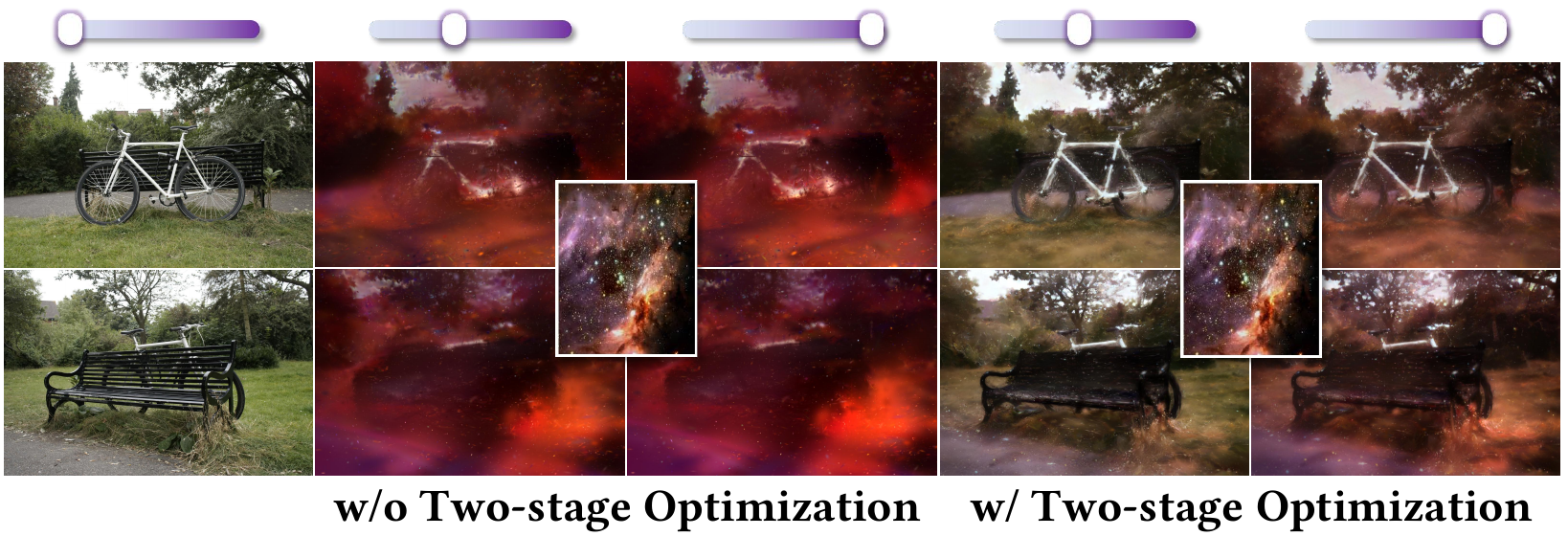}
\vspace{-5mm}
\caption{
\textbf{Results of the ablation study on two-stage optimization for tunable stylization.}
Best viewed with zoom-in.
}
\vspace{-5mm}
\label{fig:ablation_two_stage}
\end{figure}

\noindent \textbf{Two-stage Optimization for Tunable Stylization.}
To verify the effectiveness of the two-stage optimization strategy for the tunable stylization guidance, we conduct an ablation study on it.
Specifically, we employ zero-style guidance and full-style guidance to simultaneously optimize the Gaussian neurons and the style tuner, with the results presented in ``w/o Two-stage Optimization'' of~\cref{fig:ablation_two_stage}.
The results demonstrate that removing the two-stage optimization not only results in low-quality stylized outcomes (\eg, over-stylization), but also compromises the style tuner's ability to adjust the intensity of style injection.
This failure can be attributed to the unstable supervision caused by the random combination of zero-style guidance and full-style guidance, which hampers the effective learning of the initial Gaussian neurons and the style tuner.
Conversely, the two-stage optimization circumvents this issue, demonstrating high-quality stylization outcomes and flexible intensity tunability.

\begin{figure}[t]
\hsize=\linewidth
\centering
\includegraphics[width=\linewidth]{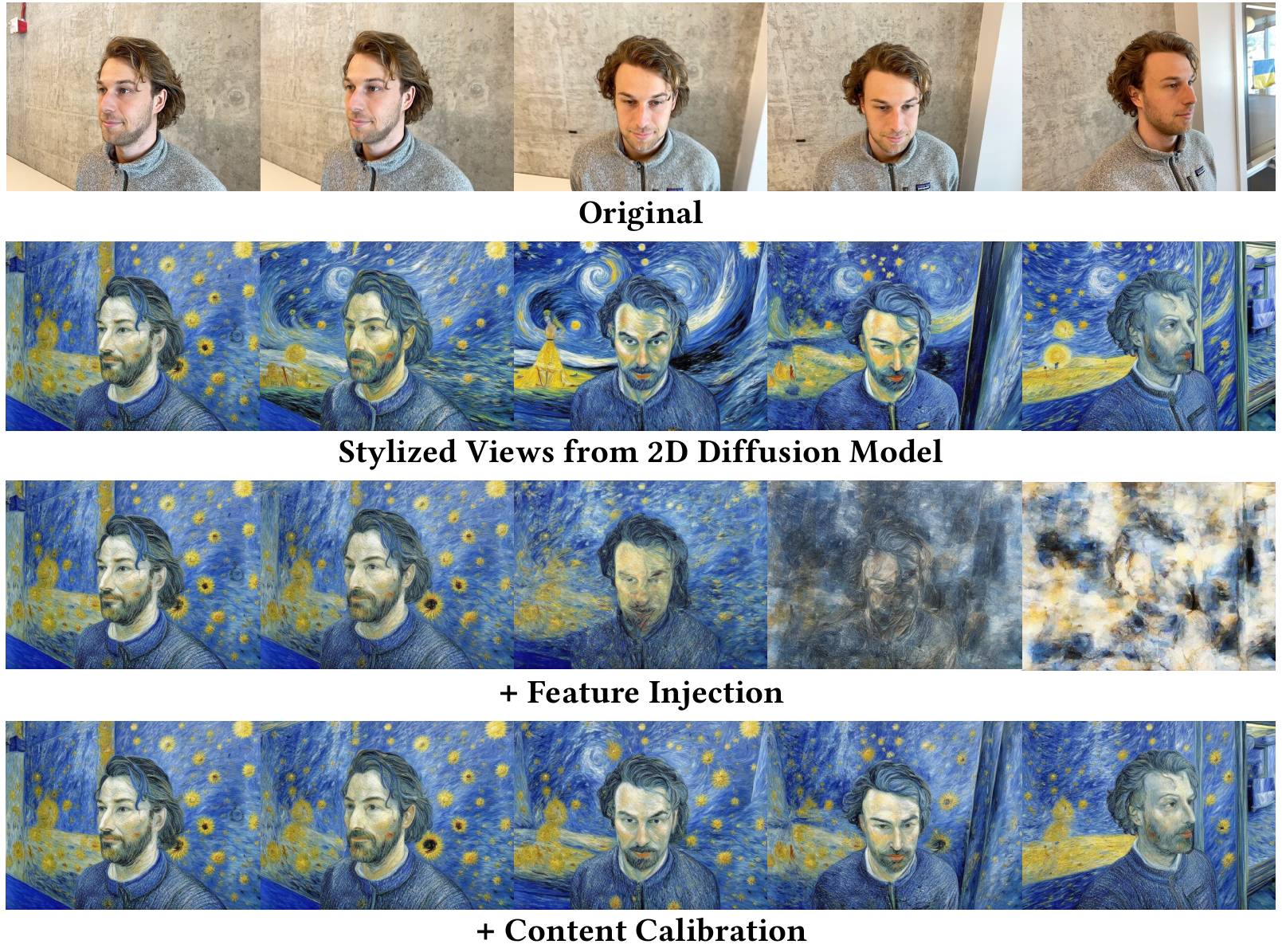}
\caption{
\textbf{Results of the ablation study on cross-view style alignment.}
Best viewed with zoom-in.
}
\vspace{-5mm}
\label{fig:ablation_mv}
\end{figure}

\noindent \textbf{Cross-view Style Alignment.}
To verify the effectiveness of the cross-view style alignment, we perform comparisons among three types of stylized views: the original stylized views, those with feature injection, and those with content calibration, as presented in~\cref{fig:ablation_mv}.
Although the original results generated from the 2D diffusion model all conform to the reference style, they lack 3D consistency, which is not conducive to optimizing 3D scenes.
After applying the feature injection, the stylized views that are close in viewpoint to the anchor view (the first column) can maintain consistent style textures.
However, as the distance from the anchor viewpoint increases, the image content of the current view tends to become distorted due to the significant differences between the injected anchor view feature and the current view feature.
After incorporating content calibration, the content and layout of the current view are preserved, while maintaining consistent style textures with the anchor view.

\section{Conclusion}
In this work, we introduce a novel intensity-tunable 3D style transfer paradigm, which enables users to flexibly adjust the style intensity injected into the scene to match their desired content-style balance, addressing the challenge posed by the existing fixed-output paradigm that struggles to adapt to diverse content-style balance requirements.
We construct an explicit modeling of style intensity and parameterize a learnable style tuner to achieve intensity-tunable style injection, and we design a tunable stylization guidance to enable efficient and stable optimization by modulating the balance between full-style guidance and zero-style guidance.
Extensive experiments demonstrate that our method not only delivers visually appealing results but also exhibits flexible customizability.
We hope that our exploration can expand the frontiers of research in 3D style transfer.

\noindent \textbf{Acknowledgements.} This work was supported in part by the Shenzhen Medical Research Funds in China (No. B2302037), National Natural Science Foundation of China (NSFC) under Grant No. 61972217, 32071459, 62176249, 62006133, 62271465, 62406167, U24B6012 and AI for Science (AI4S)-Preferred Program, Peking University Shenzhen Graduate School, China.

{
    \small
    \bibliographystyle{ieeenat_fullname}
    \bibliography{main}
}

\clearpage
\setcounter{page}{1}
\maketitleappendix

\renewcommand{\thefootnote}{\fnsymbol{footnote}}
\renewcommand{\thesection}{\Alph{section}}
\renewcommand{\thetable}{\Alph{table}}
\renewcommand{\theequation}{\Alph{equation}}
\renewcommand{\thefigure}{\Alph{figure}}

\setcounter{section}{0}
\setcounter{table}{0}
\setcounter{section}{0}
\setcounter{figure}{0}
\setcounter{equation}{0}

This Appendix contains the following parts:

\begin{itemize}
    \item \textbf{Additional Qualitative Comparison} (Appendix~\ref{sec:qualitative}). We provide the qualitative comparison with NeRF-based methods. 
    \item \textbf{Additional Visualization Results} (Appendix~\ref{sec:visualization}). We provide additional visualization results for the intensity-tunable style injection.
    \item \textbf{Details of Quantitative Evalution} (Appendix~\ref{sec:quantitative}). We provide details of the quantitative evaluation, including detailed data for calculating the CLIP metrics and evaluation criteria for the user study.
    \item \textbf{Preliminary: 3D Gaussian Splatting} (Appendix~\ref{sec:preliminary}). We introduce the technical details of 3D Gaussian Splatting.
    \item \textbf{Limitations and Social Impact} (Appendix~\ref{sec:impact}). We discuss the limitations and social impact of our work.
\end{itemize}

\section{Additional Qualitative Comparison}
\label{sec:qualitative}
\begin{figure*}[ht]
\hsize=\linewidth
\centering
\includegraphics[width=\linewidth]{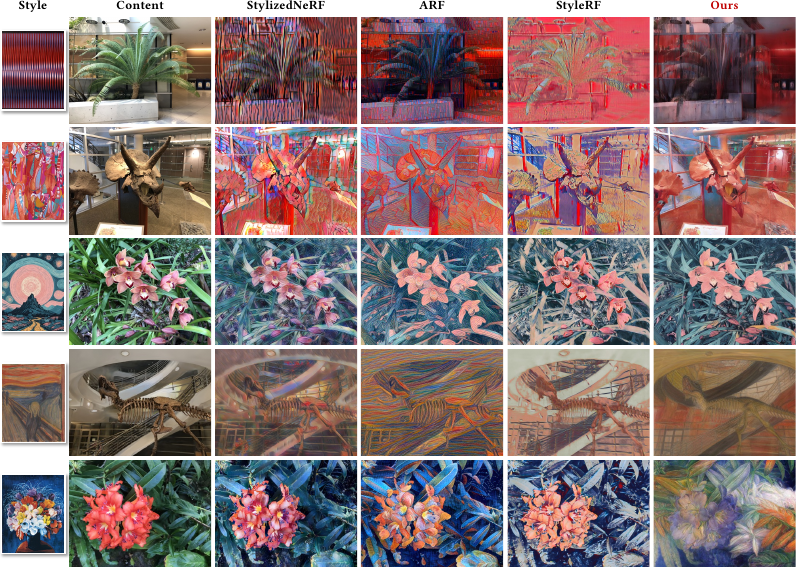}
\caption{
\textbf{Additional qualitative comparison with NeRF-based style transfer methods.} Best viewed with zoom-in.
}
\label{fig:comparison_nerf_llff}
\end{figure*}

We conduct additional qualitative comparison with existing NeRF-based style transfer methods, and the results are presented in~\cref{fig:comparison_nerf_llff}.
Concretely, We compare our method with StylizedNeRF~\cite{huang2022stylizednerf}, ARF~\cite{zhang2022arf}, and StyleRF~\cite{liu2023stylerf} on five scenes from the LLFF dataset: fern, horns, orchids, trex, flower, following their official settings.
These NeRF-based methods adhere to a fixed-output paradigm, which in some cases fails to maintain a reasonable balance between content and style, leading to suboptimal stylization results.
In contrast, our method not only produces visually appealing results, but also enables the user to adjust the style intensity injected into the scene to achieve a desired content-style balance, significantly enhancing the practicality of 3D style transfer.

\section{Additional Visualization Results}
\label{sec:visualization}
\begin{figure*}[ht]
\hsize=\linewidth
\centering
\includegraphics[width=\linewidth]{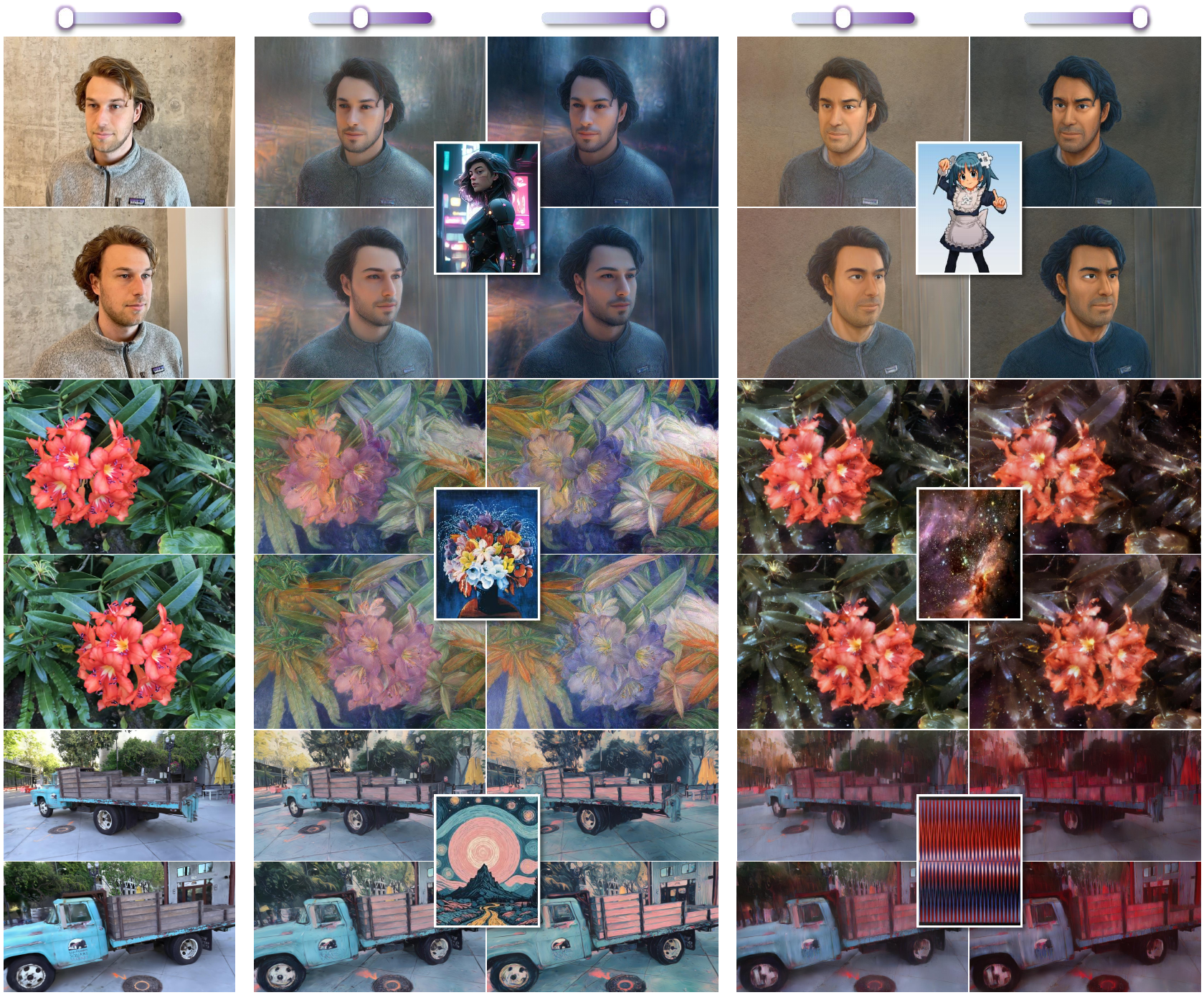}
\caption{
\textbf{More visualization results of intensity-tunable style injection.}
}
\label{fig:supp_injection}
\end{figure*}

We present more visualization results of intensity-tunable style injection, as presented in~\cref{fig:supp_injection}.
Additional results with more reference style images and 3D scenes further demonstrate the effectiveness and distinct advantages of the proposed method.

\section{Details of Quantitative Evalution}
\label{sec:quantitative}
\subsection{Evaluation Data}
We present the detailed data used for the quantitative evaluation of CLIP similarity and CLIP direction similarity in~\cref{tab:data}.
Each data sample contains the scene, the original scene description, the stylized scene description, and the reference style image.

\begin{table*}[!h]
\renewcommand{\arraystretch}{1.25}
\setlength{\tabcolsep}{9.5pt}
\begin{tabular*}{\linewidth}{>{\centering\arraybackslash}m{1cm} | >{\centering\arraybackslash}m{5cm} | >{\centering\arraybackslash}m{5cm} | >{\centering\arraybackslash}m{4cm}}
\noalign{\hrule height 1.2pt}
\textbf{Scene} & \textbf{Original Scene Description} & \textbf{Stylized Scene Description} & \textbf{Reference Style Image} \\
\hline
Face & \makecell[c]{``A man with curly hair \\in a grey jacket''} & \makecell[c]{``a Mona Lisa-style portrait of \\ a man with curly hair''} & \includegraphics[width=1.3cm]{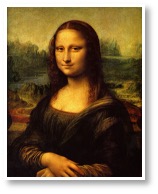} \\
\noalign{\hrule height 0.9pt}
Truck & \makecell[c]{``A vintage blue pickup truck\\ with a wooden cabin.''} & \makecell[c]{``A vintage pickup truck \\ with a wooden cabin featuring \\ vertical dark red stripes.''} & \includegraphics[width=1.3cm]{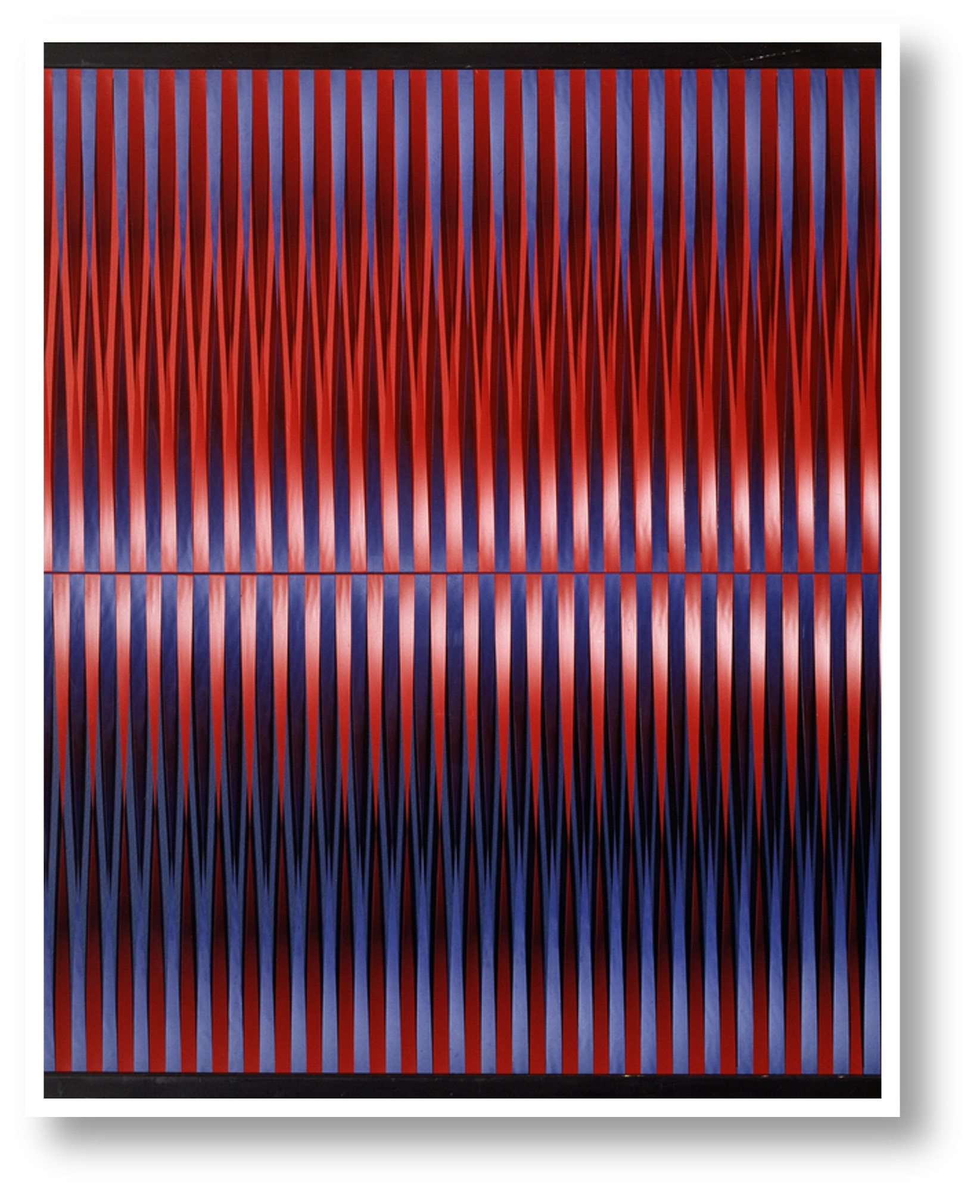} \\
\noalign{\hrule height 0.9pt}
Train & \makecell[c]{``A green Western Pacific train \\ on the tracks.''} & \makecell[c]{``An oil painting of a Western Pacific \\ train on the tracks.''} & \includegraphics[width=1.3cm]{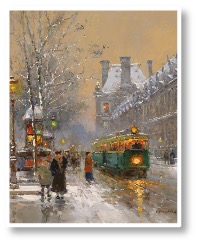} \\
\noalign{\hrule height 0.9pt}
Lego & \makecell[c]{``A Lego model of a \\ yellow bulldozer.''} & \makecell[c]{``A Lego model of a bulldozer with \\ an abstract style in shades of pink.''} & \includegraphics[width=1.3cm]{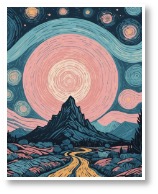} \\
\noalign{\hrule height 0.9pt}
Garden & \makecell[c]{``A garden scene with a\\ round table holding a vase.''} & \makecell[c]{``A cosmic-style depiction of a\\ garden with a round table and vase.''} & \includegraphics[width=1.3cm]{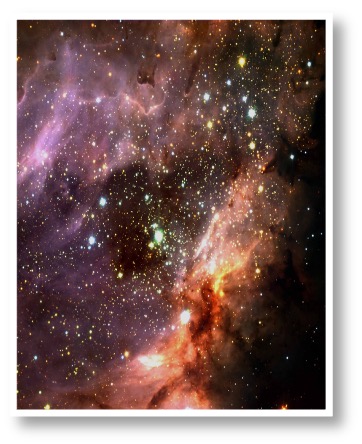} \\
\noalign{\hrule height 1.2pt}
\end{tabular*}
\caption{\textbf{Detailed data for calculating CLIP similarity and CLIP directional similarity.}}
\label{tab:data}
\end{table*}

\subsection{Evaluation Criteria for User Study}
We also present the detailed evaluation criteria for the user study.
Specifically, we request participants to evaluate from three dimensions: content consistency, style consistency, and visual appeal, using a 5-point scale. The criteria for each rating level are elaborated in~\cref{tab:study}.
We calculate the average score across the three dimensions and provide the 95\% confidence interval.
The user study results are collected from a total of $20$ participants.

\begin{table*}[!h]
\renewcommand{\arraystretch}{1.25}
\setlength{\tabcolsep}{4pt}
\begin{tabular*}{\linewidth}{>{\centering\arraybackslash}m{1.8cm} | >{\centering\arraybackslash}m{0.9cm} | m{13.9cm}}
\noalign{\hrule height 1.2pt}
\textbf{Dimension} & \textbf{\#Point} & \makecell[c]{\textbf{Description}} \\
\hline
\multirow{5}{*}{\makecell[c]{\textbf{Content} \\ \textbf{Consistency}}} & 1 & Very poor, the content is entirely unrecognizable, with severe distortion of texture and geometry. \\
\cline{2-3}
& 2 & Rather poor, the content is almost unrecognizable, with obvious distortion of texture and geometry. \\
\cline{2-3}
& 3 & Acceptable, the content is basically recognizable, with some distortion of texture and geometry. \\
\cline{2-3}
& 4 & Fairly good, the content is clearly recognizable, with minor distortion of texture and geometry. \\
\cline{2-3}
& 5 & Very good, the content is perfectly recognizable, with almost no distortion of texture and geometry. \\
\noalign{\hrule height 0.9pt}
\multirow{5}{*}{\makecell[c]{\textbf{Style} \\ \textbf{Consistency}}} & 1 & Very poor, the color and artistic style are completely inconsistent with the reference image. \\
\cline{2-3}
& 2 & Rather poor, the color and artistic style are significantly inconsistent with the reference image.  \\
\cline{2-3}
& 3 & Acceptable, the color and artistic style are basically consistent with the reference image.  \\
\cline{2-3}
& 4 & Fairly good, the color and artistic style are highly consistent with the reference image.  \\
\cline{2-3}
& 5 & Very good, the color and artistic style are completely consistent with the reference image. \\
\noalign{\hrule height 0.9pt}
\multirow{5}{*}{\makecell[c]{\textbf{Visual} \\ \textbf{Appeal}}} & 1 &  Very poor, severe artifacts, very blurry textures, overall visual effect is very poor.  \\
\cline{2-3}
& 2 & Rather poor, numerous artifacts, relatively blurry textures, overall visual effect is not satisfactory.  \\
\cline{2-3}
& 3 & Acceptable, moderate artifacts, basically clear textures, overall visual effect is basically satisfactory. \\
\cline{2-3}
& 4 & Fairly good, fewer artifacts, relatively clear textures, overall visual effect is satisfactory. \\
\cline{2-3}
& 5 & Very good, minimal artifacts, clear textures, overall visual effect is very satisfactory. \\
\noalign{\hrule height 1.2pt}
\end{tabular*}
\caption{\textbf{Detailed evaluation criteria for the user study.}}
\label{tab:study}
\end{table*}

\section{Preliminary: 3D Gaussian Splatting}
\label{sec:preliminary}
Gaussian Splatting(3DGS)~\cite{kerbl20233d} introduces splatting-based rasterization and explicitly models a scene as a collection of 3D Gaussian primitives.
Each Gaussian primitive $\Theta_i$ is parameterized by a center point $\boldsymbol{x}$ and a covariance matrix $\boldsymbol{\Sigma}_i$, which represents the Gaussian distribution as:
\begin{equation}
\label{equ:gs}
    \Theta_i (\boldsymbol{x}) = e^{-\frac{1}{2} \textit{$\boldsymbol{x}$}^{T} \boldsymbol{\Sigma}_i^{-1} \textit{$\boldsymbol{x}$}}.
\end{equation}
For the derivation of a physically meaningful covariance matrix, the following equivalent representation is applied:
\begin{equation}
    \boldsymbol{\Sigma}_i = \boldsymbol{R}_i\boldsymbol{S}_i\boldsymbol{S}_i^{T}\boldsymbol{R}_i^{T},
\end{equation}
where the covariance matrix $\boldsymbol{\Sigma}_i$ is decomposed into a scaling factor $\boldsymbol{S}_i$ and a rotation quaternion $\boldsymbol{R}_i$.
Moreover, each primitive is assigned an opacity $\sigma_i$ and a color $\boldsymbol{c}_i$ to represent the appearance of the scene.
In summary, 3DGS represents a scene as a set of Gaussian primitives: $ \Theta = \{ (\boldsymbol{\mu}_i, \boldsymbol{S}_i, \boldsymbol{R}_i, \sigma_i, \boldsymbol{c}_i) \}_{i=1}^N $, where $N$ indicates the number of primitives, and $\boldsymbol{\mu}_i$ represents the position of the center point.
In practice, 3D Gaussians can be rendered in real time to compute the color $\boldsymbol{C}$ of each pixel within the camera planes by blending $N$ ordered Gaussians overlapping the pixel:
\begin{equation}
\label{equ:volume_render}
    \boldsymbol{C} = \sum_{i \in N} \boldsymbol{c}_i \alpha_i \prod_{j=1}^{i-1} (1-\alpha_j),
\end{equation}
where $\alpha_i$ is calculated by evaluating $\Theta_i$ with~\cref{equ:gs} multiplied by its opacity $\sigma_i$.

\section{Limitations and Social Impact}
\label{sec:impact}
\subsection{Limitations}
The style transfer capability of this work is primarily limited by the underlying 2D diffusion model. Compared to 3D style transfer, image style transfer evidently has a higher success rate and more reliable results due to the larger-scale training and lower task difficulty. Therefore, despite being constrained by the image style transfer model, our method still features a higher upper bound than those 3D style transfer methods that do not leverage diffusion-based stylization priors. Additionally, our method is incapable of independently adjusting the intensity of texture style and geometric style within the stylized scene.
We will further explore intensity-tunable 3D style transfer with decoupled texture and geometry in the future.

\subsection{Social Impact}
In this work, all the 3D scenes and reference style images we utilized are publicly available.
However, the potential outcomes of our method may lead to negative societal impacts if used for harmful purposes, including intellectual property concerns for protected works and misuse of the deceptive results generated.
Consequently, we strongly advocate that users exercise responsibility when utilizing our method and strictly observe relevant ethical guidelines and legal regulations.

\end{document}